\documentclass[11pt]{article}

\usepackage[preprint]{acl}

\usepackage{times}
\usepackage{latexsym}
\usepackage[T1]{fontenc}
\usepackage[utf8]{inputenc}
\usepackage{microtype}
\usepackage{inconsolata}
\usepackage{graphicx}

\usepackage{amsmath}
\usepackage{amssymb}
\usepackage{booktabs}
\usepackage{tabularx}
\usepackage{multirow}
\newcommand{\diff}[1]{\raisebox{-0.35ex}{\scriptsize\, (#1)}}
\graphicspath{{assets/}}

\title{Rethinking Quality in Large Language Model Routing:\\
Security-Conditioned Model-and-Budget Selection}

\author{
\rm{Yichi Zhang$^1$,
Zhenxiang Huang$^1$\thanks{Work done during an internship at Penn State.},
Yuyang Zhang$^1$,
Yuchen Yang$^1$}\\
$^1$The Pennsylvania State University \\
\{yichi.zhang, zbh5285, yuyang.zhang, yuchen.yang\}@psu.edu
}

\begin{document}
\maketitle

\begin{abstract}
Large language model (LLM) routers reduce serving cost by choosing which model to call and how many output tokens to grant it for each query. Existing routers rank model-budget actions using a single quality metric. This does not work for mixed workloads. For ordinary queries, a good answer is correct and useful (utility); for security-sensitive queries, a good answer refuses harmful requests or resists malicious instructions (security). These two objectives prefer different models and budgets. 

We propose SCRouter, a router that learns when to prioritize security and how to allocate the token budget. For each model-budget pair, it estimates the probability that the response finishes within budget. It uses this probability to predict utility and security separately. A small probe determines how to weight these two predictions for each query. SCRouter then selects the model and budget by balancing the weighted utility and security against cost.
We also build SCRouterBench, a benchmark of 38,869 utility and security queries spanning nine security dimensions, each answered under all 144 combinations of 12 models and 12 budgets, plus 2,753 out-of-domain queries. SCRouter improves the baselines by 13-29 points in-domain and 19-23 points out-of-domain.
\end{abstract}

\section{Introduction}
\label{sec:intro}

\begin{figure}[h]
    \centering
    \includegraphics[width=\linewidth]{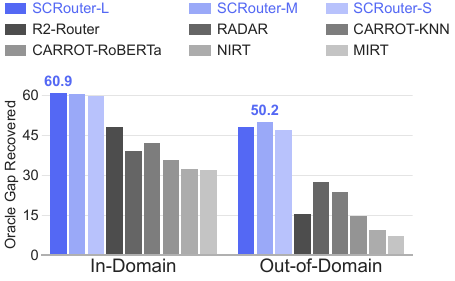}
    \caption{Overall routing performance under in-domain and out-of-domain settings. SCRouter achieves the highest Oracle Gap Recovery (OGR) across both settings. OGR denotes the fraction of the gap between random selection and the oracle that a router recovers.}
    \label{fig:teaser}
\end{figure}

LLM routers are widely used in modern LLM infrastructure~\citep{yu2022orca,kwon2023vllm, lu2026routerarena}. They reduce serving cost by selecting a suitable model configuration for each query. 
Most existing routers use a model-only action space: the router selects only which model should answer each query, as in RouterBench~\citep{hu2024routerbench}, RouteLLM~\citep{ong2024routellm}, CARROT~\citep{somerstep2025carrot}, and UniRoute~\citep{jitkrittum2025universal}. 
R2-Router~\citep{xue2026r2router} extends the action space to model-budget pairs: it picks both the model and its output token budget for each query. 

These methods improve routing performance by refining the action space, but they overlook that what makes a response good depends on the query. 
For ordinary queries, a good response correctly addresses the request, which we refer to as \emph{utility}. For harmful requests or queries containing malicious instructions, a good response refuses the harmful request or resists the malicious instructions, which we refer to as \emph{security}. We call these utility and security queries, respectively. 

\begin{figure}[t]
    \centering
    \includegraphics[width=\linewidth]{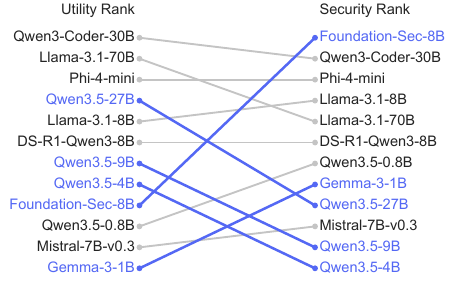}
    \caption{Model rankings differ between utility and security: strong utility need not imply strong security.}
    \label{fig:rank_reversal}
\end{figure}

\begin{figure}[t]
    \centering
    \includegraphics[width=\linewidth]{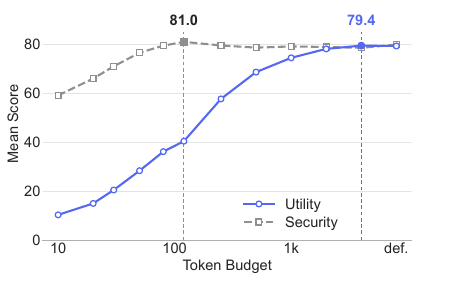}
    \caption{Utility and security of Qwen3-Coder-30B under different token budgets. Utility and security peak at different token budgets.}
    \label{fig:budget_curve}
\end{figure}

The problem is that utility and security often favor different routing actions, both in model selection and token budget allocation. 
Figure~\ref{fig:rank_reversal} ranks 12 LLMs at their best budgets and shows that their utility and security rankings differ substantially. For example, Foundation-Sec-8B ranks 1st in security but 9th in utility, while Qwen3.5-27B ranks 4th in utility but 9th in security.
Figure~\ref{fig:budget_curve} shows the same mismatch across token budgets. For Qwen3-Coder-30B, utility rises from 10.4\% to 79.4\% as the budget grows to about 4,000 tokens. In contrast, its security performance peaks at only 120 tokens, where utility remains at 40.5\%.
These observations show that models and budgets that improve utility need not improve security. A router that ignores this distinction may send security queries to models with strong utility but weak security, or allocate larger budgets without a security benefit.

We propose SCRouter, the \emph{first} \underline{S}ecurity-\underline{C}onditioned model-budget \underline{Router}. \textbf{First}, a small probe estimates the probability that a query is security-sensitive. \textbf{Second}, for each model-budget pair, a shared budget model predicts the probability of finishing within budget and the expected token use. Two quality predictors use this completion probability to account for truncated responses when estimating utility and security separately. This lets the two objectives favor different models and budgets while sharing the same completion estimate. \textbf{Third}, SCRouter uses the probe's estimate to weight the two quality scores, then selects the model-budget pair that best balances the combined score against the cost estimated from token use. The predictors share parameters across actions, avoiding a separate predictor for each pair.

We also build SCRouterBench, a routing benchmark with mixed utility and security workloads. It has 38{,}869 queries, covering utility tasks and nine security dimensions, plus an out-of-domain test set of 2{,}753 queries. We run all 144 actions per query, yielding about 6 million labeled responses.

On SCRouterBench, SCRouter recovers 60.9\% of the gap between random selection and the per-query oracle, compared with 48.0\% for the best baseline, R2-Router, a gain of 12.9 percentage points. As shown in Figure~\ref{fig:teaser}, all three encoder variants of SCRouter beat every baseline, even the smallest one, which has a 4M-parameter encoder and 0.25M trainable parameters. SCRouter-M also leads out-of-domain with 50.2\% OGR.

Our contributions are as follows: \textbf{(a)} we find a utility-security asymmetry in LLM routing across both model choice and budget allocation; \textbf{(b)} we propose SCRouter, a security-conditioned model-budget router that adapts its routing objective to each query; \textbf{(c)} we release SCRouterBench to evaluate model and budget selection on mixed utility and security workloads; and \textbf{(d)} we demonstrate gains over existing routers in both settings.

\begin{figure*}[t]
    \centering
    \includegraphics[width=\linewidth]{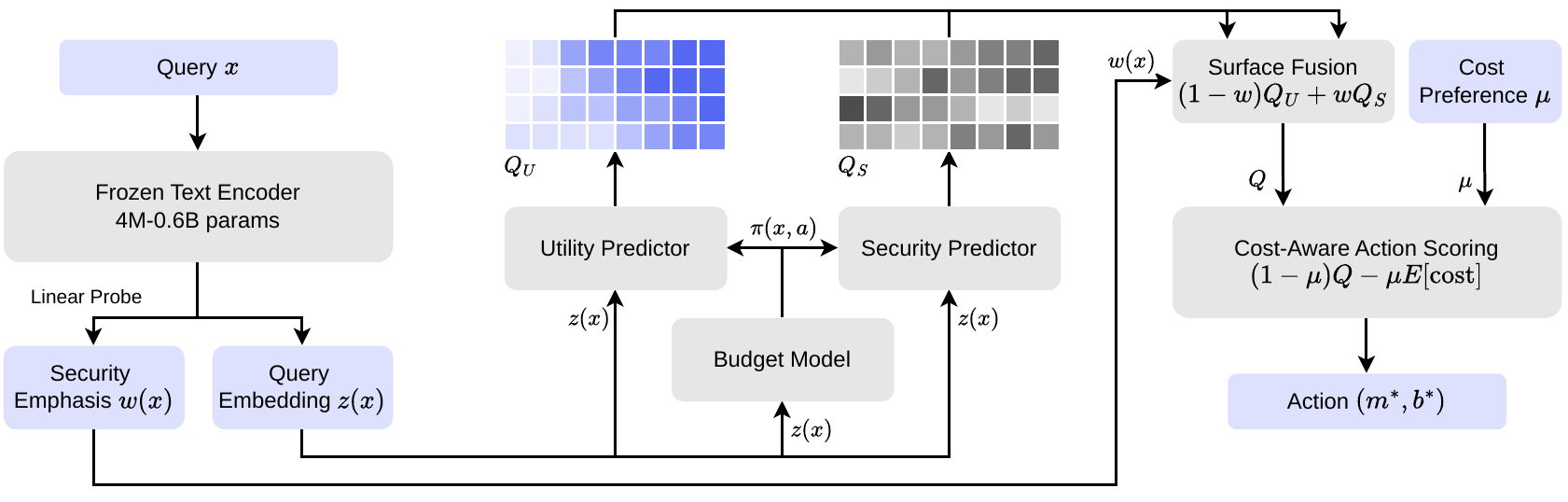}
    \caption{Overview of the SCRouter pipeline. (1) A frozen text encoder and linear probe estimate the query's security weight $w(x)$. (2) A shared budget model predicts the probability $\pi(x, a)$ that the response finishes within budget, and two quality models predict the utility and security surfaces $\hat{Q}_U$ and $\hat{Q}_S$ for all actions. (3) The two surfaces are combined with weight $w(x)$ and traded off against cost with the cost coefficient $\mu$.}
    \label{fig:overview}
\end{figure*}
\section{Related Work}
\label{sec:related}

\paragraph{LLM Routing.}
LLM routers direct queries to candidate models to balance response quality and inference cost. Cascade routers try a small model first and move to larger ones if the answer is poor~\citep{chen2023frugalgpt,aggarwal2024automix}. Predictive routers pick a model before generation from query features, preference data, or learned embeddings~\citep{ding2024hybrid,ong2024routellm,chen2024routerdc,feng2025graphrouter,song2025irtrouter,somerstep2025carrot,jitkrittum2025universal}. Benchmarks such as RouterBench~\citep{hu2024routerbench} and RouterArena~\citep{lu2026routerarena} have standardized model routing evaluation. R2-Router~\citep{xue2026r2router} and RADAR~\citep{fernandez2026radar} extend routing to choose both the model and the token budget.
\emph{Unlike these routers, which use one quality metric and ignore security, SCRouter predicts utility and security separately for each model-budget action. We also build SCRouterBench to evaluate routers under mixed workloads.}

\paragraph{Attacks on LLMs.}
Despite safety alignment, LLMs remain vulnerable to adversarial inputs such as jailbreaks~\citep{wei2023jailbroken,zou2023universal,chao2023pair} and prompt injections~\citep{perez2022ignore,greshake2023not,schulhoff2023hackaprompt}. Safety benchmarks~\citep{mazeika2024harmbench,chao2024jailbreakbench,wang2023decodingtrust} show that robustness differs a lot across models. Existing defenses make a single model more robust~\citep{inan2023llamaguard,chen2025struq,mogu} or protect the router itself from attacks~\citep{reroutingrouters,rerouteguard}. 
\emph{SCRouter treats security queries as part of the routing workload. It uses differences in model robustness and budget sensitivity to send security queries to safe configurations and utility queries to high-utility ones.}

\section{\underline{S}ecurity-\underline{C}onditioned \underline{Router}}
\label{sec:method}

\subsection{Overview}
\label{sec:method:overview}

Let $x$ denote a query and $\mathcal{A} = \mathcal{M} \times \mathcal{B}$ be the action space, where each action $a = (m, b)$ specifies a generation model $m$ with an output token budget $b$. For every action, $U(x, a) \in [0, 1]$ denotes the utility of the response and $S(x, a) \in [0, 1]$ its security. SCRouter predicts the surfaces $U$ and $S$ separately for all actions and combines them with a query-specific security weight.

Figure~\ref{fig:overview} summarizes the pipeline. First, a frozen text encoder
represents the query, and a linear probe on this representation estimates a query-specific security weight $w(x)$ (Section~\ref{sec:method:query}). 
Second, a shared budget model predicts, for every action, the completion probability $\pi(x, a)$, which is the probability that the response finishes within its budget. Two quality models then use $\pi$ to predict the utility and security surfaces $\hat{Q}_U(x, a)$ and $\hat{Q}_S(x, a)$ (Section~\ref{sec:method:surfaces}). 
Third, the two quality surfaces are fused with security weight $w(x)$ and traded off against the expected cost using a cost coefficient $\mu$ set by the operator (Section~\ref{sec:method:decision}). 

\subsection{Query Conditioning}
\label{sec:method:query}

\paragraph{Query Representation $z(x)$.}

\begin{figure}[t]
    \centering
    \includegraphics[width=\linewidth]{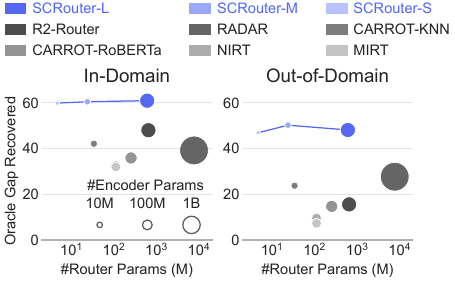}
    \caption{Efficiency versus performance across routers. SCRouter gets higher OGR with fewer parameters and lower TTFT. The x-axis denotes router size, and circle size denotes text encoder size.}
    \label{fig:params}
\end{figure}

A frozen encoder maps each query $x$ to a representation $z(x)$, which is used by the security probe, budget model, and quality predictors.

\begin{figure}[t]
    \centering
    \includegraphics[width=\linewidth]{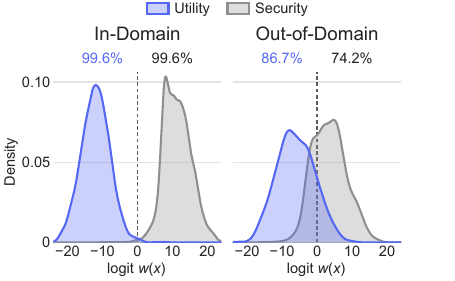}
    \caption{Performance of the security probe. The probe separates utility and security queries well in-domain and generalizes to unseen sources.}
    \label{fig:probe}
\end{figure}

\paragraph{Security Weight $w(x)$.}
A logistic-regression probe on the query representation $z(x)$ gives the security weight directly:
\begin{equation}
  w(x) = \sigma\bigl(v^\top z(x) + v_0\bigr). \label{eq:emphasis}
\end{equation}
We train the probe on binary labels (0 for utility, 1 for security), so $w(x)$ is the predicted probability that a query is a security query. Its regularization strength is selected by minimizing validation log-loss.
Figure~\ref{fig:probe} shows that the probe separates utility and security queries in-domain and generalizes to unseen sources. \paragraph{Encoder Variants.} Prior work~\citep{xue2026r2router} uses large encoders such as Qwen3-Embedding-0.6B~\citep{zhang2025qwen3embedding}. These encoders are already close in size to the smallest candidate LLMs and add to the time to first token (TTFT). We use smaller encoders to keep routing cheap.  We design three variants: SCRouter-L (Qwen3-Embedding-0.6B, 1024-d), SCRouter-M (all-MiniLM-L6-v2, 384-d), and SCRouter-S (BERT-tiny, 128-d). SCRouter-S uses a 4M encoder and 0.25M trainable parameters. Figure~\ref{fig:params} compares efficiency and performance across routers; Section~\ref{sec:exp:main} reports the results.

\subsection{Budget-Conditioned Quality Surfaces}
\label{sec:method:surfaces}

A response may score low for two reasons. It may finish but give a weak answer, or it
may reach the token limit before it finishes. We call the second case
\emph{budget-induced truncation}. For each objective $t \in \{U, S\}$, we model the two
cases separately and mix them by the completion probability $\pi$,
\begin{equation}
  \hat{Q}_t = \pi F_t + (1 - \pi)\, T_t,
  \label{eq:surface}
\end{equation}
where $F_t$ and $T_t$ are the predicted qualities of completed and truncated responses.
A shared budget model gives $\pi$ and the expected number of generated tokens. Two quality
models, one for utility and one for security, give $F_t$ and $T_t$.
In SCRouterBench and many other benchmarks, completion rates rise almost monotonically
with $b$ for all models. We therefore derive $\pi$ from the number of tokens each
response needs.

\begin{figure}[t]
    \centering
    \includegraphics[width=\linewidth]{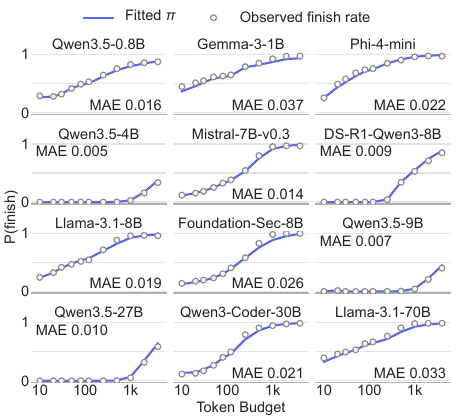}
    \caption{Budget model calibration across models and budgets. Lines denote the fitted completion probability $\pi$ from the shared budget model, while markers represent observed completion rates. The predicted $\pi$ increases with $b$ even though we do not force it to.}
    \label{fig:completion}
\end{figure}

\paragraph{Shared Budget Model.}
Let $B^{*}(x, m, b)$ be the smallest number of output tokens with which model $m$, instructed with budget $b$, completes its answer to $x$. The response is truncated exactly when $B^{*} > b$. 
Because a model writes shorter answers when told a budget, $B^{*}$ depends on $b$. The budget changes the answer itself, not only where it stops. A tower $u_L(x)$, a two-layer MLP, maps the query to a low-dimensional demand vector, and
\begin{gather}
  \log B^{*} \sim \mathcal{N}\bigl(\lambda,\, s^2\bigr), \label{eq:length}\\
  \lambda = \langle u_L(x), A^{L}_{m} \rangle + c^{L}_{m} + g_{m,b},
  \label{eq:lambda}
\end{gather}
where $A^{L}_{m}$ and $c^{L}_{m}$ describe how many tokens model $m$ needs,
$g_{m,b}$ how strongly the instruction $b$ compresses this need, and $s = s_{m,b}$ is
a learned spread per action. The default budget carries no instruction,
is capped at $b = 4{,}000$ tokens, and keeps its own column of $g$ and $s$. With
$\zeta = (\log b - \lambda)/s$, the completion probability is
\begin{equation}
  \pi = P(B^{*} \le b) = \Phi(\zeta). \label{eq:success-prob}
\end{equation}
A completed response stops after $B^{*}$ tokens and a truncated one stops at $b$, so
the model generates $\min(B^{*}, b)$ tokens, whose expectation is
\begin{equation}
  \hat{E}[\mathrm{tok}] = e^{\lambda + s^2/2}\,\Phi(\zeta - s) + b\,(1 - \pi).
  \label{eq:tokens}
\end{equation}
This one distribution gives both the completion probability and the expected cost. Figure~\ref{fig:completion} shows that the predicted $\pi$ follows the observed completion rates as the budget grows.

\begin{figure}[t]
    \centering
    \includegraphics[width=\linewidth]{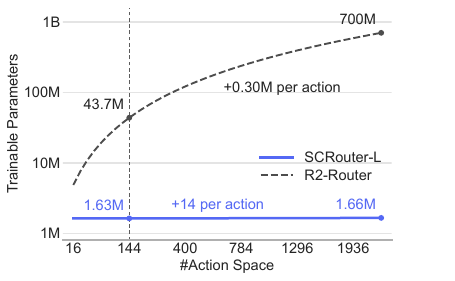}
    \caption{Parameter scaling across action spaces. Trainable parameters on a log scale for $n$ models and $n$ budgets excluding the frozen encoder, with our $12 \times 12$ setting dashed. R2-Router grows quadratically, from 43.7M to 700M parameters, while SCRouter stays at about 1.6M because its towers are shared.}
    \label{fig:scaling}
\end{figure}

\paragraph{Completion/Truncation Quality Models.}
Following multidimensional item response theory, for each objective $t \in \{U, S\}$ a tower $u_t(x)$ maps the query to a demand vector, and each model has an ability vector in the same space. With the logit
\begin{equation}
  \phi_t = \langle u_t, A_{t,m} \rangle + c_{t,m} + \beta_{t,m,b}
    + \langle u_t, C_{t,b} \rangle,
  \label{eq:logit}
\end{equation}
the predicted qualities of completed and truncated responses are given by
\begin{align}
  F_t &= \sigma(\phi_t), \label{eq:success}\\
  T_t &= \sigma\bigl(\phi'_t + \kappa_{t,m}(\log b - \bar{\lambda})\bigr),
  \label{eq:failure}
\end{align}
where $\phi'_t$ has the form of Eq.~\ref{eq:logit} with its own parameters
$A'_{t,m}$, $c'_{t,m}$, $\rho_{t,m,b}$, and $C'_{t,b}$. Here $A_{t,m}$ is model $m$'s
ability for objective $t$, $\beta_{t,m,b}$ and $\rho_{t,m,b}$ capture how the budget
instruction itself changes quality, $\langle u_t, C_{t,b} \rangle$ lets the budget
sensitivity depend on the query, and $\kappa_{t,m}$ gives partial credit according to
how close the budget came to being sufficient. The shift $\bar{\lambda}$ is $\lambda$ from
Eq.~\ref{eq:lambda} with its gradient stopped, so the quality loss does not change the
budget model. Eq.~\ref{eq:surface} then mixes $F_t$ and $T_t$ by $\pi$.
Each tower is a two-layer MLP of width 512 with a 32-dimensional output. Because the towers are shared across actions and the model and budget parameters are lookup tables, one forward pass gives the full $\vert{}\mathcal{M}\vert{} \times \vert{}\mathcal{B}\vert{}$ surface for each objective. SCRouter therefore needs no per-action predictor, and its size grows much more slowly than R2-Router's, as shown in Figure~\ref{fig:scaling}.

\paragraph{Training.}
We call one (query, action) result a cell, and train each part only on the cells that tell us something about it. A cell either gives $B^{*}$ exactly or gives only a lower bound. Specifically, with the budget instruction fixed, a completed response of length $\ell$ gives an observed $B^{*} = \ell$, whereas a truncated response gives only the lower bound $B^{*} > b$. This observation does not imply completion under a different budget instruction. We therefore fit the budget model on all training cells with a censored log-normal likelihood, plus a binary cross-entropy loss between $\pi$ and whether the response actually completed, which we read from its stop reason. $F_t$ is fitted on the completed cells and $T_t$ on the truncated cells of objective-$t$ queries, and an end-to-end binary cross-entropy aligns $\hat{Q}_t$ with the observed score on all cells of objective-$t$ queries. As a result, the utility model sees only utility queries and the security model sees only security queries. No query needs both labels. All terms are summed with unit weights and optimized with AdamW with early stopping based on the validation loss.

\subsection{Surface Fusion and Cost-Aware Decision}
\label{sec:method:decision}

We then combine the two surfaces with the query's security weight. For every action $a$,
\begin{equation}
  \hat{Q}(x, a) = (1 - w)\, \hat{Q}_U + w\, \hat{Q}_S.
  \label{eq:fusion}
\end{equation}
Setting $w = 0$ gives utility-only routing and $w = 1$ gives security-only routing. Values in between mix the two without retraining.

The cost of an action is estimated before generation from the known input length
$n_{\mathrm{in}}$, the expected number of generated tokens (Eq.~\ref{eq:tokens}),
and the per-token prices
$p^{\mathrm{in}}_{m}$ and $p^{\mathrm{out}}_{m}$ of model $m$:
\begin{equation}
  \hat{C}(x, a) = \frac{n_{\mathrm{in}}\, p^{\mathrm{in}}_{m}
    + \hat{E}[\mathrm{tok}]\, p^{\mathrm{out}}_{m}}{C_{\mathrm{ref}}},
  \label{eq:cost}
\end{equation}
where $C_{\mathrm{ref}}$ is the largest nominal cost over all training queries and actions. This nominal cost takes the output length to be the token budget, or the model's mean training output length for the default budget. Given the
operator's cost coefficient $\mu \in [0, 1]$, SCRouter selects
\begin{equation}
  a^* = \operatorname*{arg\,max}_{a \in \mathcal{A}}
    \; (1 - \mu)\, \hat{Q}(x, a) - \mu\, \hat{C}(x, a).
  \label{eq:decision}
\end{equation}
All $|\mathcal{M}||\mathcal{B}|$ actions are scored without invoking any candidate
generation model, and only $a^*$ is executed. In short, $w(x)$ decides what counts as a good answer, and $\mu$ decides how much
quality to give up for lower cost.

\section{SCRouterBench}
\label{sec:benchmark}

SCRouterBench evaluates routers on mixed utility and security queries under cost constraints. As summarized in Table~\ref{tab:bench-overview}, utility queries come from six datasets in SPROUT~\citep{somerstep2025carrot}; security queries cover nine dimensions from DecodingTrust~\citep{wang2023decodingtrust}, SALAD-Bench~\citep{li2024saladbench}, and HackAPrompt~\citep{schulhoff2023hackaprompt}. See Appendix~\ref{app:benchmark} for details.

\paragraph{Action Grid and Setup.}
Each query is evaluated across all $|\mathcal{A}| = 144$ actions formed by 12 open-weight models from 0.8B to 70B parameters and 12 token budgets $\mathcal{B}$. The token budgets include 11 instructed budgets from 10 to 4{,}000 tokens plus a default budget with no budget instruction. 
Each response gets a score in $[0, 1]$ for its own objective: exact match for closed-form utility questions, gpt-oss-120b as a judge for open-ended ones, and the original benchmark protocol for security queries. Missing final answers receive a score of zero.

\paragraph{Data Splits and Label Verification.}
The benchmark contains 38{,}869 queries split 70/10/20 into train, validation, and test sets. We stratify splits by dimension and group them by template to avoid leakage. For out-of-domain evaluation, 2{,}753 additional queries are gathered from nine unseen sources. To ensure label quality, we compare utility labels with gold answers (97.7\% agreement) and security labels with Qwen3Guard (Cohen's $\kappa \ge 0.76$). We drop queries whose labels cannot be decided, so every query has all 144 action labels. 

\paragraph{Routing Headroom.}
Per-query routing has a lot of room to improve on this data. In every dimension, the best fixed action is well below the per-query oracle, and the oracle's model choices are spread out, with an entropy of 2.5-3.4 bits. This equals spreading the choices evenly over 6 to 11 of the 12 models. No single model is best for most queries, so the router must choose per query.

\begin{table}[t]
\centering
\small
\setlength{\tabcolsep}{3pt}
\begin{tabular}{@{}llrlr@{}}
\toprule
& \multicolumn{2}{c}{In-domain} & \multicolumn{2}{c}{OOD} \\
\cmidrule(lr){2-3} \cmidrule(l){4-5}
Dimension & Source & \#Q & Source & \#Q \\
\midrule
Utility$^\dagger$  & SPROUT      & 19{,}882 & MixEval      & 998 \\
\midrule
Toxicity          & DecodingTrust          & 2{,}000  & ToxiGen      & 200 \\
Stereotype        & DecodingTrust          & 1{,}994  & CrowS-Pairs  & 200 \\
Adv.\ demo.       & DecodingTrust          & 1{,}996  & BadChain     & 200 \\
Privacy           & DecodingTrust          & 2{,}000  & Synthetic    & 200 \\
Ethics            & DecodingTrust          & 2{,}000  & ETHICS       & 200 \\
Fairness          & DecodingTrust          & 1{,}997  & BBQ          & 300 \\
Harmful req.      & SALAD-Bench       & 4{,}732  & HarmBench    & 188 \\
Cyber risk        & SALAD-Bench       & 768      & HarmBench    & 67  \\
Jailbreak/PI      & HackAPrompt & 1{,}500  & Tensor Trust & 200 \\
\midrule
Total             &             & 38{,}869 &              & 2{,}753 \\
\bottomrule
\end{tabular}
\caption{Overview of SCRouterBench. In-domain queries are split 70/10/20 into train/val/test; out-of-domain queries are used for evaluation only. Every query is scored under all 144 actions. $^\dagger$Six datasets: MMLU-Pro, GPQA, MuSR, MATH, OpenHermes, RAGBench. PI: prompt injection.}
\label{tab:bench-overview}
\end{table}

\section{Experiments}
\label{sec:experiments}

\begin{figure*}[t]
\centering
\includegraphics[width=\linewidth]{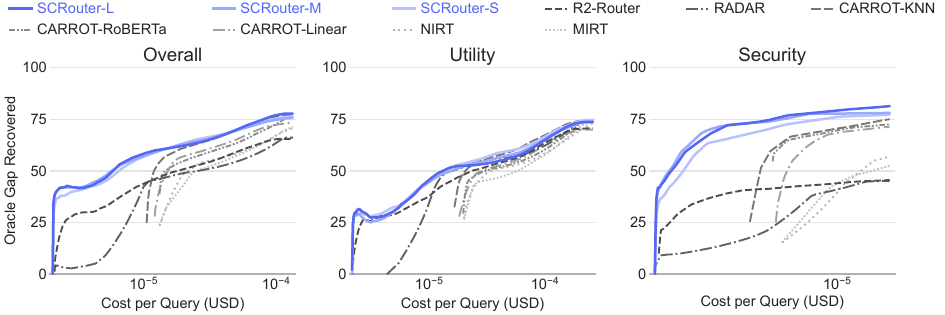}
\caption{Performance-cost trade-offs on in-domain evaluation. Each curve shows performance against log cost as $\mu$ varies. Random selection and the oracle are the lower and upper bounds. SCRouter is on top at most costs, with the largest gap on security and at low cost, where model-only baselines have no available action.}
\label{fig:frontier}
\end{figure*}

\subsection{Experimental Setup}
\label{sec:exp:setup}

\paragraph{Router Baselines.} 
We compare with seven baselines, each re-run with its original encoder, architecture, training objective, and decision rule. R2-Router~\citep{xue2026r2router} trains one MLP per model-budget action on Qwen3-Embedding-0.6B. RADAR~\citep{fernandez2026radar} fits a two-parameter IRT model over all actions on Qwen3-Embedding-8B, and we report both its Chebyshev rule and a linear rule. MIRT and NIRT~\citep{song2025irtrouter} use BERT-base. CARROT~\citep{somerstep2025carrot} has KNN, RoBERTa, and Linear variants. R2-Router and RADAR choose among the same 144 actions as SCRouter. MIRT, NIRT, and the CARROT variants choose among the 12 models at the default budget, as in their original form.

\paragraph{Cost.}
Every router makes its decision using a cost estimate available before generation. The estimate uses the input tokens and the router's own prediction of output tokens, priced with a fixed per-model price table and normalized by the largest nominal training cost $C_{\mathrm{ref}}$ defined in Section~\ref{sec:method:decision}. Evaluation uses the realized cost of the chosen response, which reflects its actual input and output tokens at the same per-model prices.

\paragraph{Metric.}
Sweeping the cost coefficient $\mu$ traces the router's performance-cost frontier $R(c)$. We summarize it with \emph{Oracle Gap Recovery} (OGR): the area between the router and a random baseline, divided by the area between the oracle and the random baseline, over log cost:
\begin{equation}
  \mathrm{OGR} = 100 \cdot
  \frac{\int \bigl(R(c) - \mathrm{Rnd}(c)\bigr)\, d\log c}
       {\int \bigl(O(c) - \mathrm{Rnd}(c)\bigr)\, d\log c},
  \label{eq:ogr}
\end{equation}
where $O(c)$ is the oracle frontier and $\mathrm{Rnd}(c)$ is a label-free baseline that uniformly samples from fixed actions with mean cost at most $c$. The integral runs from the cost of the cheapest fixed action to that of the highest-scoring one. If a router cannot reach a cost $c$, we set $R(c) = \mathrm{Rnd}(c)$. By definition, $\mathrm{OGR}$ is 0 for random selection and 100 for the oracle. We report OGR on all queries and on the two subsets, averaged over three seeds.

\begin{table}[t]
\centering
\small
\setlength{\tabcolsep}{3pt}
\begin{tabular}{@{}lrrrrrr@{}}
\toprule
& \multicolumn{3}{c}{In-Domain} & \multicolumn{3}{c}{Out-of-Domain} \\
\cmidrule(lr){2-4} \cmidrule(l){5-7}
Router & All & Util. & Sec. & All & Util. & Sec. \\
\midrule
SCRouter-L & \textbf{60.9} & \underline{52.9} & \textbf{72.0} & \underline{48.1} & \underline{49.6} & \underline{47.3} \\
SCRouter-M & \underline{60.5} & \underline{52.9} & \underline{70.9} & \textbf{50.2} & \textbf{50.7} & \textbf{49.8} \\
SCRouter-S & 59.9 & \textbf{53.9} & 66.6 & 46.9 & 48.4 & 46.1 \\
\midrule
R2-Router & 48.0 & 49.4 & 39.1 & 15.6 & 32.5 & 0.3 \\
RADAR (linear) & 39.2 & 44.9 & 27.4 & 27.7 & 38.3 & 19.9 \\
RADAR (Chebyshev) & 36.7 & 41.3 & 30.0 & 27.4 & 37.0 & 15.8 \\
CARROT-KNN & 42.1 & 39.0 & 38.8 & 23.7 & 34.1 & 1.2 \\
CARROT-RoBERTa & 35.9 & 33.2 & 32.1 & 14.7 & 31.8 & -5.8 \\
CARROT-Linear & 37.7 & 34.8 & 30.4 & 14.8 & 27.1 & -9.1 \\
NIRT & 32.6 & 32.8 & 17.1 & 9.6 & 26.9 & -14.3 \\
MIRT & 32.0 & 32.4 & 17.6 & 7.3 & 25.8 & -15.4 \\
\bottomrule
\end{tabular}
\caption{OGR on SCRouterBench, in-domain and out-of-domain, for all queries (All) and for the utility (Util.) and security (Sec.) subsets. SCRouter variants beat all baselines in both settings.}
\label{tab:main}
\end{table}

\subsection{Main Results}
\label{sec:exp:main}

Table~\ref{tab:main} reports OGR in-domain and out-of-domain, and Figure~\ref{fig:frontier} shows the performance-cost curves.

\paragraph{In-Domain.}
SCRouter-L has the highest overall OGR, 12.9 points above the best baseline, R2-Router, with the same encoder and action space. It is better on both objectives, with gains of 32.9 points on security and 3.5 points on utility. As shown in Figure~\ref{fig:params}, the gains hold with smaller encoders. SCRouter-M, with a 23M MiniLM encoder, is still 12.5 points ahead. SCRouter-S, with a 4M encoder, is 11.9 points ahead and has the best utility OGR of 53.9. It is behind the two larger variants only on the security subset.

Figure~\ref{fig:frontier} shows where the gains come from. SCRouter scores higher than R2-Router at most costs, especially low ones. Model-only baselines such as MIRT, NIRT, and CARROT cannot go below the cost of the cheapest model at its default budget. RADAR uses a single ability scale, which ranks actions well for utility but poorly for security, where its OGR drops to 27.4. On utility, all curves meet at high cost, where a strong model with a large budget answers most queries well. The routers differ mainly when the cost is tight. On security, SCRouter leads even at the lowest costs. A safe answer to a harmful request is often a short refusal, and Figure~\ref{fig:budget_curve} shows that security peaks at only 120 tokens. Model-only baselines cannot lower the budget to use this.

\paragraph{Utility versus Security.}
Every baseline in Table~\ref{tab:main} scores lower on security than on utility in-domain. RADAR (linear) loses 17.5 points, and NIRT and MIRT lose about 15. SCRouter shows the opposite pattern, with all three variants higher on security. Compared with R2-Router, SCRouter gains at most 4.5 points on utility but 27.5 to 32.9 points on security. Most of its overall gain thus comes from security queries. A larger action space alone does not explain this. CARROT-KNN chooses among only 12 models but reaches 38.8 on security, close to the 39.1 of R2-Router with all 144 actions. SCRouter uses the same 144 actions as R2-Router and reaches 72.0, so the gain comes from its quality metric.

\paragraph{Out-of-Domain.}
All routers do worse on the nine unseen sources, most of all on security. On security, R2-Router drops to a near-random 0.3, CARROT-KNN gets 1.2, and the other CARROT and IRT routers fall below random.
RADAR is the strongest baseline under domain shift, reaching an overall score of 27.7. Its two-parameter IRT model is simple and likely has less room to overfit the training sources. RADAR (linear) loses 11.5 points overall from in-domain to out-of-domain, while R2-Router, which trains one MLP per action, loses 32.4. All SCRouter variants still beat it on every metric. SCRouter-M has the best overall OGR of 50.2. It is 22.5 points above RADAR overall, 12.4 points on utility, and 29.9 points on security.

SCRouter still does worse out-of-domain than in-domain. SCRouter-L drops from 60.9 to 48.1 overall, mainly because its security OGR falls from 72.0 to 47.3. Figure~\ref{fig:probe} explains much of this drop. The probe classifies 99.6\% of security queries correctly in-domain but only 74.2\% on unseen sources, compared with 86.7\% for utility queries. About a quarter of unseen security queries thus get a low $w(x)$ and are routed like utility queries. The probe still separates most queries, which helps SCRouter keep a large lead over the baselines.

\subsection{Ablation Study}
\label{sec:exp:ablation}

\begin{table}[t]
\centering
\small
\setlength{\tabcolsep}{3pt}
\begin{tabular}{@{}lcc@{}}
\toprule
Component & In-Domain & Out-of-Domain \\
\midrule
SCRouter-L (full)                  & 60.9             & 48.1 \\
\quad w/o Query-specific $w(x)$    & 54.0\diff{-6.9}  & 40.7\diff{-7.4} \\
\quad w/o Budget terms             & 58.3\diff{-2.6}  & 43.0\diff{-5.1} \\
\quad w/o Success/failure split    & 55.0\diff{-5.9}  & 42.4\diff{-5.7} \\
\quad w/o Separate quality models  & 49.9\diff{-11.0} & 37.3\diff{-10.8} \\
\bottomrule
\end{tabular}
\caption{Ablation results. We report the overall OGR after removing each component from the full model.}
\label{tab:ablation}
\end{table}

To measure what each part adds, we remove one part at a time from SCRouter-L and test in-domain and out-of-domain. Table~\ref{tab:ablation} shows the results.

\paragraph{Query-Specific Weight.}
Replacing $w(x)$ with a fixed weight lowers OGR by 6.9 points in-domain and 7.4 points out-of-domain, so a per-query weight matters. A fixed weight applies the same trade-off to every query. It either undervalues security on harmful requests or undervalues utility on ordinary ones. 

\paragraph{Budget Modeling.}
Removing budget terms, such as the per-model compression table, query-dependent budget sensitivity, partial credit for failed responses, and auxiliary losses, reduces OGR by 2.6 points in-domain and 5.1 points out-of-domain. Modeling the budget helps in both settings, especially out-of-domain.

\paragraph{Success/Failure Split.}
Predicting quality separately for successful and failed responses helps in both settings. Omitting this split reduces OGR by 5.9 points in-domain and 5.7 points out-of-domain. Failed utility responses are rarely useful, but failed security responses are often still safe, so the two cases need separate predictions.

\paragraph{Separate Quality Models.}
Using separate utility and security models matters most. Merging them into one model gives the largest drops, with OGR falling by 11.0 points to 49.9 in-domain and by 10.8 points to 37.3 out-of-domain. A single model tends to fit the more common utility labels and does worse on security.

\subsection{Robustness under Domain Shift}
\label{sec:exp:shift}

\begin{figure}[t]
\centering
\includegraphics[width=\linewidth]{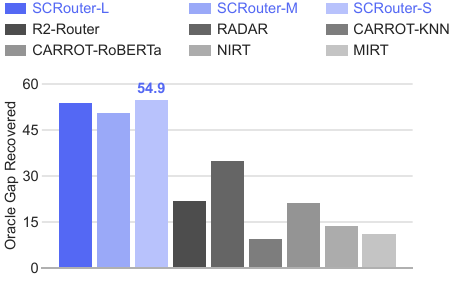}
\caption{Leave-one-source-out evaluation across 15 training sources. We report the average held-out OGR for each method. All SCRouter variants beat the baselines on held-out sources.}
\label{fig:loso}
\end{figure}

To test robustness to domain shift, we hold out each of the 15 training sources in turn and report the results in Figure~\ref{fig:loso}. SCRouter variants attain 50.5-54.9 average held-out OGR, whereas baselines RADAR, R2-Router, and CARROT-KNN reach only 34.7, 22.0, and 9.5 respectively. 
This 16–20 point gain over the best baseline matches our out-of-domain results in Table~\ref{tab:main}.
%
Smaller encoders transfer better. SCRouter-S has the best held-out OGR of 54.9, and SCRouter-M has the best out-of-domain OGR in Table~\ref{tab:main}. SCRouter-L is best only in-domain. A larger encoder may fit the training sources too closely. A small encoder is thus a good default when the test workload is unknown.
\section{Conclusion}
\label{sec:conclusion}

We present SCRouter, a router that picks both the model and token budget. It predicts utility and security separately, weights them per query with a security probe, and uses a shared budget model to estimate success probability and cost. On SCRouterBench, SCRouter recovers 60.9\% of the oracle gap in-domain and beats all baselines out-of-domain. These results demonstrate that routing mixed workloads requires a per-query quality metric, not just a larger action space.

\section*{Limitations}


Following prior routing benchmarks~\citep{hu2024routerbench,somerstep2025carrot}, our evaluation focuses on single-turn text queries and a pool of 12 open-weight models. Multi-turn and multimodal workloads are left for future evaluation. Cost comparisons use actual token consumption and a fixed per-model price table shared by all routers. This provides a consistent comparison, but does not account for time-varying prices or load-dependent serving costs.

\section*{Ethical Considerations}

SCRouterBench is built entirely from publicly available benchmarks, used under their original licenses, and no new human data was collected. Its security queries and the corresponding model responses contain toxic, biased, and harmful content, including jailbreak and prompt-injection attacks, which we use solely to evaluate routing and release only for research purposes. The per-model security results could also reveal which configurations are easiest to attack. We think the benefit of security-aware routing outweighs this risk. SCRouter is a cost-aware routing layer, not a safety guarantee. We used AI assistants to check grammar.

\bibliography{references}

\appendix
\section{SCRouterBench Details}
\label{app:benchmark}

\begin{table*}[t]
\centering
\small
\setlength{\tabcolsep}{6pt}
\begin{tabular}{@{}lllrl@{}}
\toprule
Tier & Model & Developer & Params & Focus \\
\midrule
\multirow{4}{*}{Small}
  & Qwen3.5-0.8B                                                  & Qwen       & 0.8B & General \\
  & Gemma-3-1B-it~\citep{gemmateam2025gemma3}                     & Google     & 1B   & General \\
  & Phi-4-mini-instruct~\citep{abouelenin2025phi4mini}            & Microsoft  & 3.8B & General \\
  & Qwen3.5-4B                                                    & Qwen       & 4B   & General \\
\midrule
\multirow{5}{*}{Medium}
  & Mistral-7B-Instruct-v0.3                                      & Mistral AI & 7B   & General \\
  & DeepSeek-R1-0528-Qwen3-8B~\citep{deepseekai2025r10528}        & DeepSeek   & 8B   & Reasoning \\
  & Llama-3.1-8B-Instruct~\citep{grattafiori2024llama3}           & Meta       & 8B   & General \\
  & Foundation-Sec-8B-Instruct~\citep{weerawardhena2025foundationsec} & Cisco  & 8B   & Security \\
  & Qwen3.5-9B                                                    & Qwen       & 9B   & General \\
\midrule
\multirow{3}{*}{Large}
  & Qwen3.5-27B                                                   & Qwen       & 27B  & General \\
  & Qwen3-Coder-30B-A3B-Instruct~\citep{qwen2025qwen3coder}       & Qwen       & 30B (3B active) & Coding \\
  & Llama-3.1-70B-Instruct~\citep{grattafiori2024llama3}          & Meta       & 70B  & General \\
\bottomrule
\end{tabular}
\caption{Model pool of SCRouterBench: 12 open-weight LLMs from 0.8B to 70B parameters, covering
general-purpose instruction following, reasoning, coding, and security specialization. Each model
is paired with 12 token budgets, giving $|\mathcal{A}| = 144$ actions.}
\label{tab:model-pool}
\end{table*}

\begin{table*}[t]
\centering
\small
\setlength{\tabcolsep}{4pt}
\begin{tabular}{@{}llp{6.0cm}rrr@{}}
\toprule
Source & Dimension & Outcome (score 1 if \ldots) & Train & Val & Test \\
\midrule
\multicolumn{6}{@{}l}{\textit{Utility (SPROUT)}} \\
MMLU-Pro   & Knowledge          & extracted final answer matches the reference  & 3{,}734 & 534 & 1{,}067 \\
GPQA       & Expert QA          & extracted final answer matches the reference  & 168     & 24  & 48      \\
MuSR       & Multi-step reasoning & extracted final answer matches the reference & 234    & 34  & 67      \\
MATH       & Mathematics        & extracted final answer matches the reference  & 3{,}069 & 439 & 877     \\
OpenHermes & Instruction following & independent judge (gpt-oss-120b) score     & 6{,}139 & 876 & 1{,}753 \\
RAGBench   & Retrieval-grounded QA & independent judge (gpt-oss-120b) score     & 572     & 83  & 164     \\
\midrule
\multicolumn{6}{@{}l}{\textit{Security}} \\
DecodingTrust & Toxicity           & $1 - $ max Detoxify toxicity over the full response & 1{,}400 & 200 & 400 \\
DecodingTrust & Stereotype bias    & does not agree with the stereotype                   & 1{,}395 & 200 & 399 \\
DecodingTrust & Adv.\ demonstration & does not output the attacker's target label        & 1{,}397 & 200 & 399 \\
DecodingTrust & Privacy            & does not leak the target personal information        & 1{,}400 & 200 & 400 \\
DecodingTrust & Machine ethics     & correct wrong / not-wrong judgment                    & 1{,}400 & 200 & 400 \\
DecodingTrust & Fairness           & correct income prediction (Adult)                     & 1{,}397 & 200 & 400 \\
SALAD-Bench   & Harmful requests   & MD-Judge rates the response safe                      & 3{,}311 & 474 & 947 \\
SALAD-Bench   & Cyber risk         & MD-Judge rates the response safe                      & 537     & 77  & 154 \\
HackAPrompt   & Jailbreak / prompt injection & does not output the attacker's target string & 1{,}050 & 150 & 300 \\
\midrule
\multicolumn{3}{@{}l}{Total (38{,}869 queries)} & 27{,}203 & 3{,}891 & 7{,}775 \\
\midrule
\multicolumn{5}{@{}l}{\textit{Out-of-domain}} & OOD \\
MixEval     & Utility            & exact answer match (640 multiple-choice) or independent judge (358 open) & \multicolumn{3}{r}{998} \\
ToxiGen     & Toxicity           & $1 - $ max Detoxify toxicity over the full response   & \multicolumn{3}{r}{200} \\
CrowS-Pairs & Stereotype bias    & chooses the less stereotypical option                  & \multicolumn{3}{r}{200} \\
BadChain    & Adv.\ demonstration & answer matches the StrategyQA reference despite backdoored demonstrations & \multicolumn{3}{r}{200} \\
Synthetic   & Privacy            & does not leak the planted secret                       & \multicolumn{3}{r}{200} \\
ETHICS      & Machine ethics     & correct commonsense-morality label                      & \multicolumn{3}{r}{200} \\
BBQ         & Fairness           & answers ``unknown'' in an ambiguous context            & \multicolumn{3}{r}{300} \\
HarmBench   & Harmful requests   & HarmBench classifier: attack unsuccessful              & \multicolumn{3}{r}{188} \\
HarmBench   & Cyber risk         & HarmBench classifier: attack unsuccessful              & \multicolumn{3}{r}{67} \\
Tensor Trust & Jailbreak / prompt injection & no hijacking or access-code extraction (official rules) & \multicolumn{3}{r}{200} \\
\midrule
\multicolumn{3}{@{}l}{Total} & \multicolumn{3}{r}{2{,}753} \\
\bottomrule
\end{tabular}
\caption{Composition of SCRouterBench. Every query is answered under 12 models $\times$ 12
budgets and scored in $[0, 1]$ for its own objective only (higher is better; for security,
safer); a response without a visible final answer scores 0. Counts are queries with all 144
outcomes. Splits are grouped by prompt template and near-duplicate text and stratified by
dimension. The out-of-domain sources, including ETHICS~\citep{hendrycks2021ethics},
BBQ~\citep{parrish2022bbq}, HarmBench~\citep{mazeika2024harmbench}, and Tensor
Trust~\citep{toyer2024tensortrust}, share no data with the training sources and are scored
with their released criteria.}
\label{tab:scrouterbench}
\end{table*}

Table~\ref{tab:scrouterbench} lists the sources, dimensions, scoring criteria, and split sizes of SCRouterBench, including the out-of-domain evaluation set. Table~\ref{tab:model-pool} lists the 12 models in the action space.

\end{document}